\documentclass[10pt,twocolumn,letterpaper]{article}

\usepackage{cvpr}                

\usepackage{cuted} 

\usepackage{currfile} 

\usepackage{caption} 

\usepackage{multirow}
\usepackage{array}

\definecolor{cvprblue}{rgb}{0.21,0.49,0.74}
\usepackage[pagebackref,breaklinks,colorlinks,allcolors=cvprblue]{hyperref}

\def\paperID{*****} 
\def\confName{CVPR}
\def\confYear{2026}

\title{EEG Decoding Using CNN and LSTM Network}

\author{Athanasios Karagounis\\
School of Science, Department of Digital Industry Technologies\\
GR34400, Psachna, Greece\\
{\tt\small akaragun@gs.uoa.gr}
}

\begin{document}
\maketitle

\begin{abstract}
Motor imagery (MI) brain--computer interfaces (BCIs) have emerged as a promising approach for establishing flexible communication pathways between the human brain and external devices , particularly for individuals affected by stroke or neurodegenerative disorders. Reliable decoding of motor-imagery electroencephalography (MI-EEG) remains challenging because EEG recordings contain substantial noise and exhibit complex, weakly informative relationships with the underlying brain activity. Although deep learning provides an effective means of learning representations directly from EEG signals, its application to MI-EEG feature learning remains comparatively limited. This study introduces a hybrid deep-learning architecture that integrates a convolutional neural network (CNN) with a bidirectional long short-term memory (bi-LSTM) network. The CNN is used to learn high-level spatial and temporal representations directly from raw MI-EEG recordings, whereas the bi-LSTM models temporal dependencies and relationships among the extracted features. The proposed approach is evaluated using both a publicly available dataset and a privately acquired dataset obtained with an EEG acquisition system. The experimental results indicate that the CNN\&bi-LSTM architecture provides robust performance for both two- and three-class motor-imagery classification and demonstrates promising subject-independent decoding capability across the evaluated methods.
\end{abstract}

\vspace{0.5em}

\section{Introduction}
\label{sec:intro}

Brain--computer interfaces (BCIs)~\cite{ref1,ref2,ref3} provide a direct communication pathway between neural activity and the external environment~\cite{ref4}. Such systems are particularly relevant when conventional peripheral neural pathways are impaired by conditions including stroke and neurodegenerative disorders. An increasingly important BCI application is the monitoring and interpretation of brain activity for recognizing user intentions~\cite{ref5}. Among the available BCI modalities, motor-imagery electroencephalography (MI-EEG)~\cite{ref6} is especially attractive because it can distinguish patterns associated with imagined movements without requiring overt physical motion. Nevertheless, MI-EEG signals are affected by multiple sources of noise and generally exhibit a low signal-to-noise ratio \cite{amanatiadis2009digital}. Consequently, accurate interpretation of EEG activity is a central requirement for achieving reliable EEG-based BCI performance.

Several conventional approaches have been widely employed for MI-EEG classification. Linear Discriminant Analysis (LDA)~\cite{ref7} seeks to express a dependent variable through a linear combination of input features. It can provide satisfactory results for approximately linear problems, but its effectiveness may decrease for nonlinear BCI data. Support Vector Machines (SVMs)~\cite{ref8} can construct nonlinear decision boundaries by mapping observations into higher-dimensional spaces through nonlinear functions, providing useful nonlinear modeling and generalization capabilities. Naive Bayes (NB)~\cite{ref9} is derived from Bayes' theorem and assumes conditional independence among features; this assumption can be problematic for highly noisy and correlated EEG measurements. Deep neural networks (DNNs)~\cite{ref10}, trained using backpropagation (BP)~\cite{ref11}, provide strong nonlinear approximation capabilities, although their optimization may be affected by vanishing gradients~\cite{ref12} and training-efficiency issues~\cite{ref13}. More importantly, deep models can learn latent structures directly from raw measurements, thereby reducing reliance on manually designed feature-selection and feature-extraction stages.

Conventional deep-learning pipelines based on individual convolutional neural networks (CNNs)~\cite{ref14,faniadis2020deep,amanatiadis2018understanding} or recurrent neural networks (RNNs)~\cite{ref15} still require substantial effort for model training, parameter optimization, and EEG feature engineering. Manual extraction and selection of features in the time and frequency domains can be computationally demanding and does not necessarily guarantee that the most discriminative features or temporal relationships will be identified. To overcome these limitations, this work combines CNNs with long short-term memory (LSTM) networks~\cite{ref16}. LSTM architectures are well suited to learning dependencies over relatively long temporal intervals and can therefore model the relationship between EEG activity and imagined movements, such as imagining a left-hand movement. CNNs are also attractive for EEG analysis because convolution and pooling operations can naturally perform signal-processing functions, including localized filtering and energy-related feature extraction, while allowing the filters to be learned directly from the data.

CNN-based processing is particularly useful for EEG because convolutional operations can learn signal transformations directly from the training data. In this framework, operations analogous to band-pass filtering and energy extraction can be represented through learned convolutional and pooling layers. This data-driven formulation reduces the need to prescribe all signal-processing characteristics in advance and allows the network to adapt its feature representation to the classification task.

\begin{figure}[t]
  \centering
  \includegraphics[width=\linewidth]{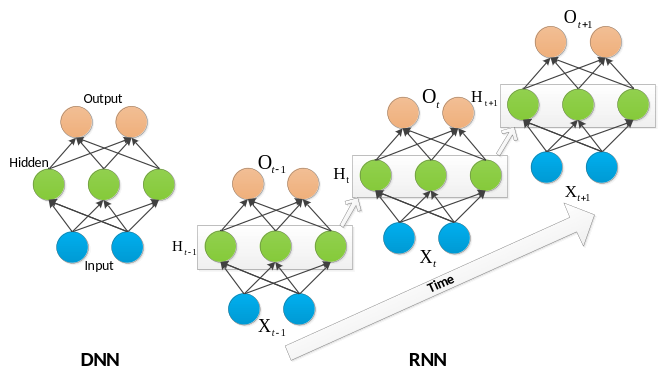}
  \caption{Schematic representation of a deep neural network and an unfolded recurrent neural network.}
  \label{fig:dnn_rnn}
\end{figure}

Recurrent neural networks (RNNs) are a class of artificial neural networks in which recurrent connections create directed cycles, as illustrated in \cref{fig:dnn_rnn}. This recurrent structure enables the network to retain information from preceding observations and is therefore suitable for sequential data. In contrast to conventional feedforward networks, where inputs are generally treated independently, RNNs explicitly account for dependencies between successive elements of a sequence. For time-series prediction, the current output can consequently depend on information extracted from earlier observations. Although RNNs are theoretically capable of representing dependencies over arbitrarily long sequences, conventional implementations often struggle to retain information over extended time intervals. Long short-term memory (LSTM) networks were introduced to alleviate the exploding- and vanishing-gradient problems associated with conventional RNN training~\cite{ref16}. They have subsequently demonstrated strong performance in a broad range of sequential-data applications. Standard LSTMs estimate the current output using preceding inputs, whereas bidirectional LSTMs exploit information propagated from both earlier and later observations~\cite{ref17}.

To obtain informative and robust EEG representations for reliable intention classification, we develop a hybrid deep-learning architecture in which a CNN performs feature extraction and a bidirectional LSTM captures temporal dynamics and higher-level feature relationships. The proposed framework, together with the associated data-processing, feature-extraction, and classification procedures, is described in detail below. A series of experiments and visual analyses are subsequently conducted to assess the effectiveness and robustness of the CNN\&bi-LSTM approach. The main contributions of this work are summarized as follows.

\begin{enumerate}
  \item The proposed CNN\&bi-LSTM architecture is designed to retain informative long-range temporal patterns while improving the classification performance of MI-EEG-based BCIs.
  \item The recognition framework learns spatial and temporal dependencies directly from raw EEG measurements through the CNN feature extractor. A bidirectional LSTM is then employed to model sequential dynamics and improve the representation of temporal dependencies.
  \item The proposed approach is extensively assessed using both a private dataset and a publicly available dataset. The experimental findings demonstrate high classification accuracy together with consistent performance across the evaluated subjects and datasets.
\end{enumerate}

\section{Dataset Acquisition and Representation}
\label{sec:dataset}

\begin{figure}[t]
  \centering
  \includegraphics[width=0.9\linewidth]{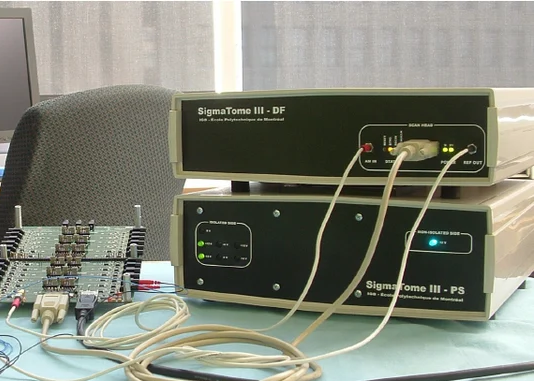}
  \caption{EEG acquisition equipment and experimental setup.}
  \label{fig:equipment}
\end{figure}

The proposed method is evaluated using two types of data: a private dataset acquired by the authors and publicly available MI-EEG datasets. The recorded sequences were organized into labeled training trials and test trials, with the principal dataset characteristics summarized in \cref{tab:datasets}. Dataset D1 corresponds to the private recordings obtained with the EEG acquisition equipment illustrated in \cref{fig:equipment}, whereas D2--D4 are publicly available datasets. During the acquisition procedure, healthy participants with a mean age of 28.7 years wore the EEG device and remained seated in front of a computer display. The screen provided instructions indicating the motor actions that participants were required to imagine.

A sampling frequency of 1000\,Hz and a recording duration of 200\,s were selected for the experimental recordings in order to maintain consistent signal lengths across subjects. The characteristics of the datasets used in the study are summarized below:
\begin{equation}
D^{j} = \left\{ \left( X_{1}^{j},y_{1}^{j} \right),\left( X_{2}^{j},y_{2}^{j} \right),\ldots,\left( X_{L}^{j},y_{L}^{j} \right) \right\},
\end{equation}
where $X_{L}^{j}$ denotes the $L$-th preprocessed trial of subject $j$ and $y_{L}^{j}$ its associated label. For each trial and subject $j$, the preprocessed signal length is denoted by $L = R \times T = 200{,}000$~\cite{ref16}. For dataset D1, the three-dimensional input tensor contains $N = 240$ trials, $L = 200{,}000$ temporal samples, and $E = 64$ EEG electrode channels~\cite{ref17}.

\begin{table}[t]
  \centering
  \small
  \setlength{\tabcolsep}{2.5pt}
  \caption{Characteristics of the raw datasets.}
  \label{tab:datasets}
  \begin{tabular}{lcccc}
    \toprule
    \multirow{2}{*}{Datasets} & Private & \multicolumn{3}{c}{Public} \\
    \cmidrule(lr){2-2} \cmidrule(lr){3-5}
     & D1 & D2 & D3 & D4 \\
    \midrule
    Subjects          & 6   & 7   & 12  & 5 \\
    Electrodes        & 64  & 64  & 32  & 64 \\
    Trials            & 240 & 128 & 278 & 200 \\
    Sample rate (Hz)  & 500 & 1000 & 1000 & 1000 \\
    Tasks             & \begin{tabular}[c]{@{}c@{}}left, right\\ hand, tongue\end{tabular}
                      & \begin{tabular}[c]{@{}c@{}}left, right\\ hand\end{tabular}
                      & \begin{tabular}[c]{@{}c@{}}left, right\\ hand\end{tabular}
                      & \begin{tabular}[c]{@{}c@{}}left, right\\ hand\end{tabular} \\
    \bottomrule
  \end{tabular}
\end{table}

\section{Proposed Architecture}
\label{sec:method}

\subsection{Data Pre-processing}

Before the EEG recordings were supplied to the deep convolutional network, a sequence of preprocessing operations was performed to improve signal quality and establish a consistent input representation.

\vspace{0.3em}
\noindent\textbf{1) Referencing~\cite{ref18}.} In this experiment, the vertex electrode (Cz) is used as the reference electrode. We construct the $M \times N$ data matrix which contains the two-dimensional information for each electrode along with the Cz-referenced EEG recordings. The process can be denoted as the following transformation:
\begin{equation}
C_{z}\left( V_{m} \right) = \left( I - V_{Cz} \right) V_{m},
\end{equation}
where $I$ is the identity matrix and
\begin{equation}
V_{Cz} = \begin{bmatrix}
0 & \cdots\ 1\ \cdots & 0 \\
\vdots & \vdots & \vdots \\
0 & \cdots\ 1\ \cdots & 0
\end{bmatrix}
\end{equation}
is an $M \times M$ matrix with ones only in the column corresponding to the Cz electrode.

\vspace{0.3em}
\noindent\textbf{2) Electrode selection.} The principal MI-EEG electrode locations considered in the processing stage include P3, P4, C3, C4, O1, O2, Pz, Fz, and Cz. The Cz electrode, used as the reference, was positioned at the central region of the scalp.

\vspace{0.3em}
\noindent\textbf{3) Noise removal.} Wavelet-based denoising was applied to reduce unwanted components and obtain cleaner EEG signals before classification.

\vspace{0.3em}
\noindent\textbf{4) Signal filtering.} The signals were filtered within the 8--23\,Hz frequency range, covering the frequency components associated with the $\mu$ and beta rhythms.

\subsection{Deep Neural Network Architecture}

\begin{figure*}[t]
  \centering
  \includegraphics[width=0.92\textwidth]{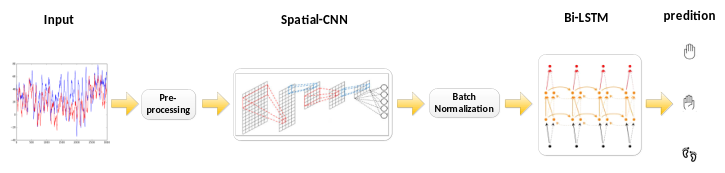}
  \caption{Overall architecture of the proposed CNN\&bi-LSTM framework, including the preprocessing stage, the spatial CNN, batch normalization, and the bidirectional LSTM classifier.}
  \label{fig:architecture}
\end{figure*}

The overall CNN\&bi-LSTM architecture is presented in \cref{fig:architecture}. The framework consists of a preprocessing stage followed by a convolutional neural network and a bidirectional LSTM module for modeling temporal dynamics in the sequential EEG representation. The CNN is a multilayer neural architecture whose parameters are optimized through error backpropagation \cite{amanatiadis2020social}. The spatial convolution stage is intended to transform the original EEG channels into a more informative, task-oriented feature representation in a common latent space. Eight spatial filters are employed in this layer, producing eight feature representations after convolution. The convolution kernel has a size of $[36 \times 1]$, and each resulting feature has a size of $(1 \times 60)$. The signal is subsequently reshaped to $(L, C, 1)$ to enable batch normalization \cite{sophokleous2022educational}, where $C$ denotes the effective number of feature channels. The spatial convolution operation therefore performs learned spatial filtering of the EEG input. Following convolution and nonlinear activation~\cite{ref19,amanatiadis2018interpolation}, the representation from the preceding layer is transformed into the corresponding output feature representation:
\begin{equation}
O_{m}^{2} = f\!\left( \sum_{j = 1}^{j \leq 36} X_{i,j} \times k_{m}^{2} + b_{m}^{2} \right),
\end{equation}
where $k_{m}^{2}$ and $X_{i,j}$ are the convolution kernel of $[28 \times 1]$ and the input, respectively, and $b_{m}^{2}$ is the bias.

Batch normalization (BN)~\cite{ref20} is incorporated to improve generalization and accelerate optimization by normalizing the network inputs. After feature extraction, the remaining EEG representation is organized according to temporal order. BN is then applied so that each feature dimension is normalized with a mean of zero and a standard deviation of one. For an input with dimensionality $L$, each feature dimension is normalized according to the corresponding batch-normalization formulation:
\begin{equation}
BN(X_{i}) = \frac{X_{i} - E\lbrack X_{i}\rbrack}{\sqrt{Var\lbrack X_{i}\rbrack}},
\end{equation}
\begin{equation}
E\left\lbrack X_{i} \right\rbrack = \frac{1}{m}\sum_{i = 1}^{L} X_{i},
\end{equation}
\begin{equation}
Var\left\lbrack X_{i} \right\rbrack = \frac{1}{m}\sum_{i = 1}^{L}\left( X_{i} - E\left\lbrack X_{i} \right\rbrack \right)^{2}.
\end{equation}

Although conventional RNN and LSTM architectures are effective for sequential modeling, their ability to represent complex dependencies occurring at multiple temporal scales can remain limited. To address this issue, the proposed recurrent component is designed to capture complementary dependencies over different timescales and to learn latent temporal structures. The bidirectional LSTM processes the sequence in both directions, thereby incorporating information from preceding as well as subsequent observations. This formulation improves the representation of temporal context and provides greater robustness during training, including against gradient-related difficulties. The network is subsequently trained and evaluated using cross-validation.

\subsection{Bidirectional LSTM Network Architecture}

\begin{figure*}[t]
  \centering
  \includegraphics[width=0.92\textwidth]{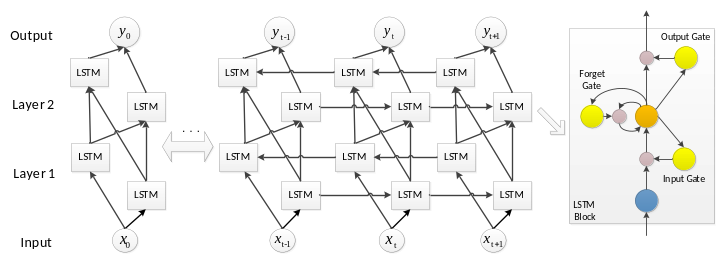}
  \caption{Structural decomposition of the proposed bidirectional LSTM network.}
  \label{fig:bilstm}
\end{figure*}

The detailed bidirectional LSTM architecture is illustrated in \cref{fig:bilstm}. Each white block represents a forward or backward unit within a hidden layer and corresponds to an LSTM cell.

The LSTM architecture was developed to mitigate the vanishing-gradient problem encountered in conventional RNNs. As shown in the right-hand part of \cref{fig:bilstm}, each LSTM block contains a memory cell together with a forget gate, input gate, and output gate. The cell maintains information over extended time intervals, which provides the memory mechanism that gives the LSTM its name.

A conventional LSTM generates its current output using information from preceding inputs, whereas a bidirectional LSTM incorporates contextual information from both earlier and later observations~\cite{ref21,ref22,ref23}. The purpose of this configuration is to exploit the complete temporal context when estimating the output by combining the forward and backward representations. In this way, information from both temporal directions can contribute to improved recognition performance. The state update of the forward layer is defined by the corresponding recurrent equations:
\begin{align}
c_{t} &= z_{t} \odot c_{t - 1} + i_{t} \odot \sigma\left( W_{c}x_{t} + U_{c}c_{t - 1} + b_{c} \right), \\
z_{t} &= \sigma\left( W_{f}x_{t} + U_{f}h_{t - 1} + b_{f} \right), \\
i_{t} &= \sigma\left( W_{i}x_{t} + U_{i}h_{t - 1} + b_{i} \right), \\
o_{t} &= \sigma\left( W_{o}x_{t} + U_{o}h_{t - 1} + b_{o} \right), \\
h_{t}^{f} &= o_{t} \odot \tanh\left( c_{t} \right).
\end{align}

The backward layer is computed using the same set of operations but in the reverse temporal direction. In the formulation, $i$, $o$, $f$ and $c$ represent the input gate, output gate, forget gate, and cell state, respectively, while the associated weight matrices $W$, $U$ and biases $b$ are optimized during training. The sigmoid function $\sigma$ is applied element-wise, and the Hadamard product $\odot$ denotes element-wise multiplication. As shown in \cref{fig:bilstm}, the LSTM therefore produces the forward hidden sequence $h_{t}^{f}$ and the backward hidden sequence $h_{t}^{b}$ together with the resulting output sequence $y_{t}$ through the recurrent update equations:
\begin{equation}
y_{t} = \upsilon\left( W_{hy}h_{t}^{f} + W_{ky}h_{t}^{b} + b_{y} \right),
\end{equation}
where $\upsilon$ denotes the output activation function.

The resulting representation is formed through the combined processing of the LSTM units described above. The bidirectional configuration provides complementary information from the two temporal directions and can therefore improve the overall representation quality. The same processing strategy is subsequently applied to the second bi-LSTM layer.

The initialization procedure begins with one unit whose parameters are independent of the remaining network structure. The coefficients obtained from LASSO are used to initialize the weight matrices $W$, allowing the hidden unit to begin from a solution corresponding to least-squares regression~\cite{ref24}. The remaining parameters are initialized randomly. Consequently, the network starts from an output representation related to least-angle regression and can then be optimized to learn temporal correlations while reducing the mean squared error~\cite{ref25,ref26}.

Mean squared error (MSE) is selected instead of cross-correlation because the filtered MI-EEG recordings contain substantial intervals corresponding to resting states. The training objective is therefore formulated as the minimization of the MSE:
\begin{equation}
\Gamma\!\left\lbrack \left\{ \left( x_{1},y_{1} \right),\ldots,\left( x_{l},y_{l} \right) \right\} \right\rbrack = \sum_{i = 1}^{l} E\!\left\lbrack \left\| \widetilde{y}_{i} - y_{i} \right\| \right\rbrack,
\end{equation}
where $y_{i}$ and $\widetilde{y}_{i}$ denote the backpropagated outputs associated with the ground-truth and predicted signals, respectively, and $\left\| \cdot \right\|$ represents the specified $L^{2}$ error measure. In addition to the conventional training procedure, supervised pre-training is employed to address the multi-task learning problem while maintaining a shared representation in the lower hidden layers. This strategy provides a more informative initialization for the subsequent task-specific training stages.

\section{Experimental Results}
\label{sec:results}

The proposed CNN\&bi-LSTM model is evaluated using both the private and public datasets. First, the learning behavior and spatial performance of the proposed architecture are compared with alternative methods. The learned feature patterns at different network layers are then examined visually. Finally, classification accuracy and training efficiency are analyzed to assess the overall effectiveness of the proposed model.

\subsection{Learning Rates and Spatial Performance}

\begin{figure}[t]
  \centering
  \begin{subfigure}{\linewidth}
    \centering
    \includegraphics[width=0.92\linewidth]{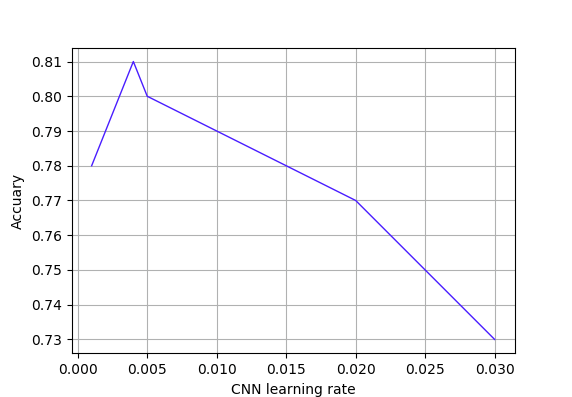}
    \caption{CNN learning rate}
    \label{fig:lr_cnn}
  \end{subfigure}
  \\[0.6em]
  \begin{subfigure}{\linewidth}
    \centering
    \includegraphics[width=0.92\linewidth]{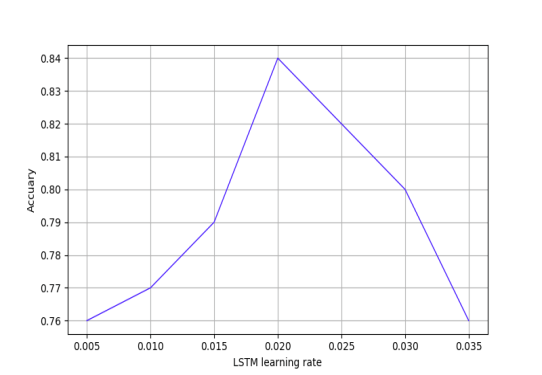}
    \caption{LSTM learning rate}
    \label{fig:lr_lstm}
  \end{subfigure}
  \caption{Comparison of learning-rate settings for the convolutional and recurrent components.}
  \label{fig:lr}
\end{figure}

The parameters of neural networks are commonly optimized through gradient descent. During each training iteration, backpropagation calculates the derivative of the loss with respect to each parameter, after which the parameters are updated accordingly. Directly applying these gradients can produce large parameter variations between consecutive iterations and may contribute to unstable optimization or overfitting. A learning-rate coefficient is therefore introduced to scale the gradient before each parameter update. \Cref{fig:lr} illustrates the behavior of the proposed components under different learning-rate settings. Based on these experiments, learning rates of 0.004 for the CNN and 0.02 for the LSTM were selected.

\begin{figure}[t]
  \centering
  \includegraphics[width=0.88\linewidth]{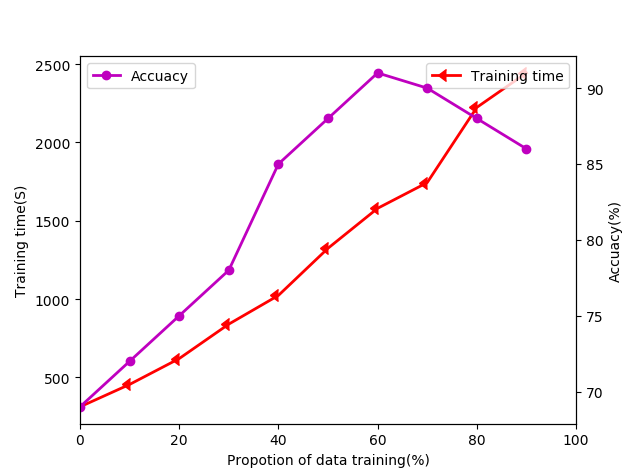}
  \caption{Classification accuracy and training time as a function of the training-data proportion.}
  \label{fig:datasplit}
\end{figure}

\Cref{fig:datasplit} illustrates how the amount of training data affects the classification performance of the proposed model. When 60\% of the available data is used for training, the model achieves approximately 89\% accuracy. Increasing the proportion of training data produces only a modest additional improvement. This observation suggests that the proposed architecture is relatively robust to reductions in the amount of training data~\cite{ref27}. The results also indicate that training time increases approximately linearly with the size of the training dataset.

\begin{figure}[t]
  \centering
  \includegraphics[width=0.88\linewidth]{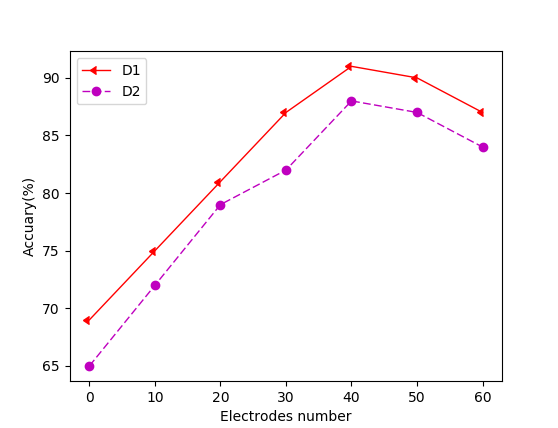}
  \caption{Spatial performance comparison between D1 and D2 using the proposed bidirectional LSTM.}
  \label{fig:spatial}
\end{figure}

The number of electrodes used during spatial feature extraction is another important parameter~\cite{ref28}. Using an excessive number of channels may increase the risk of overfitting and estimation errors, whereas using too few channels may discard useful discriminative information. \Cref{fig:spatial} compares the classification performance obtained with different numbers of electrodes. The best performance was obtained when 40 electrodes were used. Increasing the number beyond this point resulted in reduced performance, which is attributed to overfitting. The D1 dataset also produced better results than D2, which may be related to external disturbances affecting the latter experimental configuration.

\subsection{EEG Feature Patterns at Different Network Layers}

\begin{figure}[t]
  \centering
  \begin{subfigure}{\linewidth}
    \centering
    \includegraphics[width=0.85\linewidth]{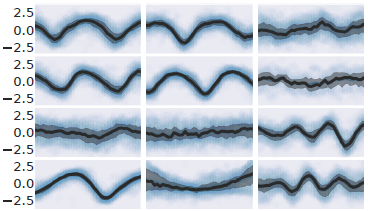}
    \caption{Layer 1}
    \label{fig:layer1}
  \end{subfigure}
  \\[0.6em]
  \begin{subfigure}{\linewidth}
    \centering
    \includegraphics[width=0.85\linewidth]{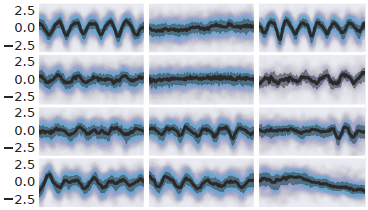}
    \caption{Layer 2}
    \label{fig:layer2}
  \end{subfigure}
  \caption{EEG representations generated at the different network layers: (a) first layer; (b) second layer.}
  \label{fig:layers}
\end{figure}

\Cref{fig:layers} presents the EEG representations produced by the first and second layers (upper and lower panels, respectively) for a randomly selected filter and each subject.

The blue points indicate the standardized scores associated with the principal activating input windows for a selected filter, while the median response is represented within the interquartile range. The median patterns in the earlier layer exhibit structures that are closer to portions of sinusoidal waveforms, whereas the representations produced by subsequent layers become progressively more complex.

\subsection{Classification Accuracy and Training Efficiency Across Subjects}

To assess the effectiveness of the proposed bidirectional LSTM architecture for motor-imagery classification, a systematic set of experiments was conducted. The proposed method was compared with LDA, SVM, CNN, and RNN using the private dataset D1 and the public datasets D2--D4. The results for D1 are reported in \cref{tab:accuracy}. Implementations of the comparison methods were developed in Python and their parameters were tuned for the experiments. The proposed architecture performed consistently across the seven subjects and achieved a mean classification accuracy of 85.4\%, compared with 69.5\% for LDA, 73.0\% for SVM, 75.9\% for CNN, and 79.5\% for RNN. Although the improvement is moderate, it demonstrates the benefit of combining convolutional and bidirectional recurrent processing. The proposed model achieved its highest subject-specific accuracy of 92.5\% for subject C, followed by 91.4\% for subject E and 90.3\% for subject F. Batch normalization also reduced training time for most of the evaluated configurations, as shown in \cref{tab:bn}.

\begin{table*}[t]
  \centering
  \small
  \setlength{\tabcolsep}{6pt}
  \caption{Recognition accuracy (\%) obtained with the evaluated methods on the private dataset D1.}
  \label{tab:accuracy}
  \begin{tabular}{lcccccccc}
    \toprule
    Method & SubA & SubB & SubC & SubD & SubE & SubF & SubG & Mean \\
    \midrule
    LDA          & 67.7 & 70.3 & 72.4 & 68.2 & 69.8 & 71.5 & 66.3 & 69.5 \\
    SVM          & 70.6 & 74.3 & 71.3 & 75.1 & 69.2 & 73.1 & 77.4 & 73.0 \\
    CNN          & 75.8 & 77.2 & 78.3 & 79.4 & 73.4 & 78.3 & 69.3 & 75.9 \\
    RNN          & 82.3 & 79.3 & 78.8 & 79.4 & 75.5 & 83.5 & 78.3 & 79.5 \\
    LSTM         & 78.3 & 82.1 & 87.7 & 75.7 & 82.4 & \textbf{90.3} & 84.5 & 83.0 \\
    CNN\&bi-LSTM & 82.3 & 84.4 & \textbf{92.5} & 78.5 & \textbf{91.4} & 86.3 & 82.4 & \textbf{85.4} \\
    \bottomrule
  \end{tabular}
\end{table*}

\begin{table*}[t]
  \centering
  \small
  \setlength{\tabcolsep}{6pt}
  \caption{Comparison of training time (hh:mm:ss) with and without batch normalization (BN).}
  \label{tab:bn}
  \begin{tabular}{llccccccc}
    \toprule
    Method & BN & SubA & SubB & SubC & SubD & SubE & SubF & SubG \\
    \midrule
    \multirow{2}{*}{\begin{tabular}[c]{@{}l@{}}1-layer\\ bi-LSTM\end{tabular}}
      & without & 00:44:37 & 00:35:31 & 00:39:14 & 00:47:52 & 00:35:23 & 00:43:19 & 00:37:53 \\
      & with    & 00:38:37 & 00:33:31 & 00:32:14 & 00:51:52 & 00:28:42 & 00:38:23 & 00:35:11 \\
    \midrule
    \multirow{2}{*}{\begin{tabular}[c]{@{}l@{}}2-layer\\ bi-LSTM\end{tabular}}
      & without & 01:11:37 & 00:38:37 & 00:55:32 & 01:04:27 & 00:41:37 & 00:51:33 & 00:48:11 \\
      & with    & 00:48:33 & 00:41:37 & 00:43:22 & 00:54:31 & 00:38:34 & 00:49:32 & 00:40:03 \\
    \bottomrule
  \end{tabular}
\end{table*}

The results obtained on the public datasets are summarized in \cref{tab:public} and compared with the corresponding alternative approaches. The CNN\&bi-LSTM model consistently provides higher accuracy than the other evaluated machine-learning and recurrent architectures. For D2, the two-layer ST-CNN and RNN models achieve 74.57\% and 77.42\%, respectively, whereas the one-layer LSTM reaches 79.56\%. For D3, ST-CNN and the two-layer RNN obtain 79.83\% and 82.43\%, respectively, while the one-layer LSTM reaches 84.78\%. For D4, the corresponding accuracies are 81.30\% for ST-CNN, 83.10\% for RNN, and 82.67\% for LSTM. Increasing the LSTM depth alone does not necessarily provide a consistent improvement; however, the proposed CNN\&bi-LSTM configuration achieves 81.80\%, 87.97\%, and 85.32\% on D2, D3, and D4, respectively. These results indicate that combining convolutional feature extraction with bidirectional temporal modeling is more effective than simply increasing the number of recurrent parameters or layers.

\begin{table}[t]
  \centering
  \small
  \caption{Classification accuracy (\%) on the public datasets.}
  \label{tab:public}
  \begin{tabular}{lccc}
    \toprule
    Method & D2 (\%) & D3 (\%) & D4 (\%) \\
    \midrule
    SVM          & 73.40 & 75.20 & 77.70 \\
    LDA          & 61.37 & 66.95 & 67.95 \\
    RF           & 76.20 & 81.16 & 80.30 \\
    ST-CNN       & 74.57 & 79.83 & 81.30 \\
    RNN          & 75.94 & 81.16 & 83.10 \\
    LSTM         & 77.42 & 82.43 & 82.67 \\
    bi-LSTM      & 79.56 & 84.78 & 83.53 \\
    CNN\&bi-LSTM & \textbf{81.80} & \textbf{87.97} & \textbf{85.32} \\
    \bottomrule
  \end{tabular}
\end{table}

\begin{figure}[t]
  \centering
  \includegraphics[width=0.88\linewidth]{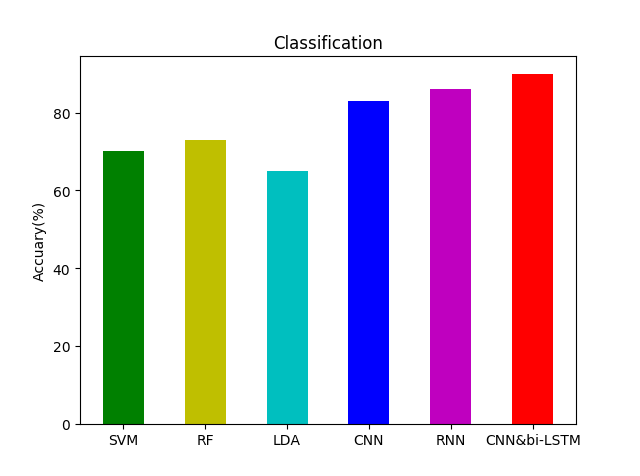}
  \caption{Accuracy comparison among the evaluated methods for the custom dataset (D1).}
  \label{fig:comparison}
\end{figure}

As illustrated in \cref{fig:comparison}, the proposed architecture achieves stronger classification performance than the alternative methods considered in the study. The improvement in recognition accuracy supports the suitability of the CNN\&bi-LSTM framework for MI-EEG decoding applications.

\section{Conclusion}
\label{sec:conclusion}

This paper presented a motor-imagery EEG recognition framework that combines convolutional neural networks, batch normalization, and a bidirectional LSTM recurrent architecture to learn spatial and temporal dependencies directly from raw EEG recordings. The spatial convolution layer acts as a learned spatial filter, while batch normalization improves the optimization process and contributes to better generalization during MI-EEG training. The extracted representations are subsequently processed by the bi-LSTM network, which integrates information from both past and future portions of the sequence and can therefore capture more complex temporal relationships. Overall, the proposed CNN\&bi-LSTM architecture provides a unified mechanism for transforming lower-level EEG measurements into higher-level representations that incorporate both temporal and frequency-related information. The experimental results demonstrate that the method can effectively emphasize informative components of MI-EEG signals and achieve competitive classification performance across the evaluated datasets and subjects. The use of bidirectional recurrent modeling therefore represents a potential alternative to conventional spectral-feature-based EEG analysis and may support future research and practical applications of MI-EEG-based rehabilitation systems.

{
    \small
    \bibliographystyle{ieeenat_fullname}
    \bibliography{main}

\begin{thebibliography}{34}
\providecommand{\natexlab}[1]{#1}
\providecommand{\url}[1]{\texttt{#1}}
\expandafter\ifx\csname urlstyle\endcsname\relax
  \providecommand{\doi}[1]{doi: #1}\else
  \providecommand{\doi}{doi: \begingroup \urlstyle{rm}\Url}\fi

\bibitem[Amanatiadis et~al.(2018{\natexlab{a}})Amanatiadis, Kaburlasos, and
  Kosmatopoulos]{amanatiadis2018interpolation}
Angelos Amanatiadis, Vasileios~G Kaburlasos, and Elias~B Kosmatopoulos.
\newblock Interpolation kernels in fully convolutional networks and their
  effect in robot vision tasks.
\newblock In \emph{2018 IEEE International Conference on Imaging Systems and
  Techniques (IST)}, pages 1--5, 2018{\natexlab{a}}.

\bibitem[Amanatiadis et~al.(2018{\natexlab{b}})Amanatiadis, Kaburlasos, and
  Kosmatopoulos]{amanatiadis2018understanding}
Angelos Amanatiadis, Vasileios~G Kaburlasos, and Elias~B Kosmatopoulos.
\newblock Understanding deep convolutional networks through gestalt theory.
\newblock In \emph{2018 IEEE international conference on imaging systems and
  techniques (IST)}, pages 1--6, 2018{\natexlab{b}}.

\bibitem[Amanatiadis et~al.(2020)Amanatiadis, Kaburlasos, Dardani,
  Chatzichristofis, and Mitropoulos]{amanatiadis2020social}
Angelos Amanatiadis, Vasileios~G Kaburlasos, Christina Dardani, Savvas~A
  Chatzichristofis, and Athanasios Mitropoulos.
\newblock Social robots in special education: Creating dynamic interactions for
  optimal experience.
\newblock \emph{IEEE Consumer Electronics Magazine}, 9\penalty0 (3):\penalty0
  39--45, 2020.

\bibitem[Amanatiadis and Andreadis(2009)]{amanatiadis2009digital}
Angelos~A Amanatiadis and Ioannis Andreadis.
\newblock Digital image stabilization by independent component analysis.
\newblock \emph{IEEE Transactions on instrumentation and measurement},
  59\penalty0 (7):\penalty0 1755--1763, 2009.

\bibitem[Belwafi et~al.(2018)Belwafi, Romain, et~al.]{ref7}
Kais Belwafi, Olivier Romain, et~al.
\newblock An embedded implementation based on adaptive filter bank for
  brain-computer interface systems.
\newblock \emph{Journal of Neuroscience Methods}, 2018.

\bibitem[Chiarelli et~al.(2018{\natexlab{a}})Chiarelli, Croce, Merla,
  et~al.]{ref3}
A.~M. Chiarelli, P. Croce, A. Merla, et~al.
\newblock Deep learning for hybrid {EEG}-{fNIRS} brain-computer interface:
  application to motor imagery classification.
\newblock \emph{Journal of Neural Engineering}, 15\penalty0 (3),
  2018{\natexlab{a}}.

\bibitem[Chiarelli et~al.(2018{\natexlab{b}})Chiarelli, Croce, et~al.]{ref10}
Antonio~Maria Chiarelli, Pierpaolo Croce, et~al.
\newblock Deep learning for hybrid {EEG}-{fNIRS} brain-computer interface:
  application to motor imagery classification.
\newblock \emph{Journal of Neural Engineering}, 15\penalty0 (3),
  2018{\natexlab{b}}.

\bibitem[Costecalde et~al.(2018)Costecalde, Aksenova, et~al.]{ref13}
Thomas Costecalde, Tetiana Aksenova, et~al.
\newblock A long-term {BCI} study with {ECoG} recordings in freely moving rats.
\newblock \emph{Neuromodulation}, 21\penalty0 (2):\penalty0 149--159, 2018.

\bibitem[Coyle and McGinnity(2010)]{ref19}
Damien Coyle and T.~Martin McGinnity.
\newblock Improving the separability of multiple {EEG} features for a {BCI} by
  neural-time-series-prediction-preprocessing.
\newblock \emph{Biomedical Signal Processing and Control}, 5\penalty0
  (3):\penalty0 196--204, 2010.

\bibitem[Emami and Chau(2018)]{ref1}
Zahra Emami and Tom Chau.
\newblock Investigating the effects of visual distractors on the performance of
  a motor imagery brain-computer interface.
\newblock \emph{Clinical Neurophysiology}, pages 1268--1275, 2018.

\bibitem[Faniadis and Amanatiadis(2020)]{faniadis2020deep}
Efstathios Faniadis and Angelos Amanatiadis.
\newblock Deep learning inference at the edge for mobile and aerial robotics.
\newblock In \emph{2020 IEEE International Symposium on Safety, Security, and
  Rescue Robotics (SSRR)}, pages 334--340, 2020.

\bibitem[Ioffe and Szegedy(2015)]{ref21}
S. Ioffe and C. Szegedy.
\newblock Batch normalization: accelerating deep network training by reducing
  internal covariate shift.
\newblock \emph{arXiv preprint arXiv:1502.03167}, 2015.

\bibitem[Li et~al.(2016)Li, Zhang, and Luo]{ref16}
Mingai Li, Meng Zhang, and Xinyong Luo.
\newblock Combined long short-term memory based network employing wavelet
  coefficients for {MI}-{EEG} recognition.
\newblock In \emph{IEEE International Conference on Mechatronics and Automation
  (ICMA)}, pages 1971--1976, 2016.

\bibitem[Li et~al.(2017{\natexlab{a}})Li, Zhu, and Zhang]{ref17}
M. Li, W. Zhu, and M. Zhang.
\newblock The novel recognition method with optimal wavelet packet and {LSTM}
  based recurrent neural network.
\newblock In \emph{IEEE International Conference on Mechatronics and Automation
  (ICMA)}, 2017{\natexlab{a}}.

\bibitem[Li et~al.(2017{\natexlab{b}})Li, Zhu, and Zhang]{ref18}
M. Li, W. Zhu, and M. Zhang.
\newblock The novel recognition method with optimal wavelet packet and {LSTM}
  based recurrent neural network.
\newblock In \emph{IEEE International Conference on Mechatronics and Automation
  (ICMA)}, pages 584--589, 2017{\natexlab{b}}.

\bibitem[Lin et~al.(2013)Lin, Chen, and Yan]{ref20}
M. Lin, Q. Chen, and S. Yan.
\newblock Network in network.
\newblock \emph{arXiv preprint arXiv:1312.4400}, 2013.

\bibitem[Lotte et~al.(2018)Lotte, Bougrain, Cichocki, et~al.]{ref2}
F. Lotte, L. Bougrain, A. Cichocki, et~al.
\newblock A review of classification algorithms for {EEG}-based brain-computer
  interfaces: a 10-year update.
\newblock \emph{Journal of Neural Engineering}, 15\penalty0 (3), 2018.

\bibitem[Lu et~al.(2017)Lu, Li, and Ren]{ref25}
Na Lu, Tengfei Li, and Xiaodong Ren.
\newblock A deep learning scheme for motor imagery classification based on
  restricted {B}oltzmann machines.
\newblock \emph{IEEE Transactions on Neural Systems and Rehabilitation
  Engineering}, 25\penalty0 (6):\penalty0 566--576, 2017.

\bibitem[Manor and Geva(2015)]{ref14}
Ran Manor and Amir~B. Geva.
\newblock Convolutional neural network for multi-category rapid serial visual
  presentation {BCI}.
\newblock \emph{Frontiers in Computational Neuroscience}, 9:\penalty0 146,
  2015.

\bibitem[Mazumder et~al.(2015)Mazumder, Rakshit, et~al.]{ref11}
Ankita Mazumder, Arnab Rakshit, et~al.
\newblock A back-propagation through time based recurrent neural network
  approach for classification of cognitive {EEG} states.
\newblock In \emph{IEEE International Conference on Engineering and Technology
  (ICETECH)}, pages 55--59, 2015.

\bibitem[Miao et~al.(2017)Miao, Zeng, and Wang]{ref9}
Minmin Miao, Hong Zeng, and Aimin Wang.
\newblock Discriminative spatial-frequency-temporal feature extraction and
  classification of motor imagery {EEG}: an sparse regression and weighted
  naive {B}ayesian classifier-based approach.
\newblock \emph{Journal of Neuroscience Methods}, 278:\penalty0 13--24, 2017.

\bibitem[Nguyen et~al.(2015)Nguyen, Nahavandi, Khosravi, Creighton, and
  Hettiarachchi]{ref5}
T. Nguyen, S. Nahavandi, A. Khosravi, D. Creighton, and I. Hettiarachchi.
\newblock {EEG} signal analysis for {BCI} application using fuzzy system.
\newblock In \emph{International Joint Conference on Neural Networks (IJCNN)},
  pages 1--8, 2015.

\bibitem[Park et~al.(2017)Park, Lee, and Sim]{ref27}
Seung-Min Park, Tae-Ju Lee, and Kwee-Bo Sim.
\newblock Heuristic feature extraction method for {BCI} with harmony search and
  discrete wavelet transform.
\newblock \emph{International Journal of Control, Automation and Systems},
  130:\penalty0 127--131, 2017.

\bibitem[Ramaswamy and Bhatnagar(2018{\natexlab{a}})]{ref12}
A. Ramaswamy and S. Bhatnagar.
\newblock Analysis of gradient descent methods with nondiminishing bounded
  errors.
\newblock \emph{IEEE Transactions on Automatic Control}, 63\penalty0
  (5):\penalty0 1465--1471, 2018{\natexlab{a}}.

\bibitem[Ramaswamy and Bhatnagar(2018{\natexlab{b}})]{ref26}
A. Ramaswamy and S. Bhatnagar.
\newblock Analysis of gradient descent methods with nondiminishing bounded
  errors.
\newblock \emph{IEEE Transactions on Automatic Control}, 63\penalty0
  (5):\penalty0 1465--1471, 2018{\natexlab{b}}.

\bibitem[Salehinejad et~al.(2017)Salehinejad, Sankar, and Barfett]{ref15}
H. Salehinejad, S. Sankar, and J. Barfett.
\newblock Recent advances in recurrent neural networks.
\newblock \emph{arXiv preprint arXiv:1801.01078}, 2017.

\bibitem[Sophokleous et~al.(2022)Sophokleous, Amanatiadis, Gkelios, and
  Chatzichristofis]{sophokleous2022educational}
Aphrodite Sophokleous, Angelos Amanatiadis, Socratis Gkelios, and Savvas~A
  Chatzichristofis.
\newblock Educational robotics in the service of the gestalt similarity
  principle.
\newblock In \emph{2022 IEEE International Conference on Consumer Electronics
  (ICCE)}, pages 1--6, 2022.

\bibitem[Sturm et~al.(2016)Sturm, Bach, Samek, and M{\"u}ller]{ref4}
I. Sturm, S. Bach, W. Samek, and K.-R. M{\"u}ller.
\newblock Interpretable deep neural networks for single-trial {EEG}
  classification.
\newblock \emph{Journal of Neuroscience Methods}, 274:\penalty0 141--145, 2016.

\bibitem[Tang et~al.(2016)Tang, Li, and Sun]{ref6}
Z.~C. Tang, C. Li, and S.~Q. Sun.
\newblock Single-trial {EEG} classification of motor imagery using deep
  convolutional neural networks.
\newblock \emph{Optik}, 130:\penalty0 11--18, 2016.

\bibitem[Thomas et~al.(2017{\natexlab{a}})Thomas, Maszczyk, and Sinha]{ref23}
J. Thomas, T. Maszczyk, and N. Sinha.
\newblock Deep learning-based classification for brain-computer interfaces.
\newblock In \emph{IEEE International Conference on Systems, Man, and
  Cybernetics (SMC)}, pages 234--239, 2017{\natexlab{a}}.

\bibitem[Thomas et~al.(2017{\natexlab{b}})Thomas, Maszczyk, Sinha,
  et~al.]{ref28}
John Thomas, Tomasz Maszczyk, Nishant Sinha, et~al.
\newblock Deep learning-based classification for brain-computer interfaces.
\newblock In \emph{IEEE International Conference on Systems, Man, and
  Cybernetics (SMC)}, pages 234--239, 2017{\natexlab{b}}.

\bibitem[Xie et~al.(2018)Xie, Schwartz, and Prasad]{ref22}
Z. Xie, O. Schwartz, and A. Prasad.
\newblock Decoding of finger trajectory from {ECoG}.
\newblock \emph{Journal of Neural Engineering}, 14:\penalty0 036009, 2018.

\bibitem[Yeh et~al.(2018)Yeh, Mokhtar, Prasad, et~al.]{ref8}
Sai~Chong Yeh, Norrima Mokhtar, Girijesh Prasad, et~al.
\newblock Automated classification and removal of {EEG} artifacts with {SVM}
  and wavelet-{ICA}.
\newblock \emph{IEEE Journal of Biomedical and Health Informatics}, 22\penalty0
  (3):\penalty0 664--670, 2018.

\bibitem[Yulita et~al.(2017)Yulita, Fanany, and Arymurthy]{ref24}
I.~N. Yulita, M.~I. Fanany, and A.~M. Arymurthy.
\newblock Combining deep belief networks and bidirectional long short-term
  memory.
\newblock In \emph{International Conference on Electrical Engineering and
  Informatics (ICEEI)}, pages 1--6, 2017.

\end{thebibliography}
}

\end{document}